\documentclass[runningheads]{llncs}
\usepackage{eccv}
\usepackage[T1]{fontenc}
\usepackage{graphicx}
\usepackage{booktabs}
\usepackage{amsmath}
\usepackage{multirow}
\usepackage{url}
\usepackage{xcolor}
\usepackage{xspace}
\usepackage{tikz}
\usetikzlibrary{arrows.meta,positioning,shapes.geometric,fit,calc,shadows}
\allowdisplaybreaks

\newcommand{\method}{CapMap-MS-TTA\xspace}
\definecolor{boxA}{RGB}{46,125,50}
\definecolor{boxB}{RGB}{21,101,192}
\definecolor{boxC}{RGB}{194,24,91}
\definecolor{boxM}{RGB}{69,90,100}

\usepackage[hidelinks]{hyperref}

\begin{document}

\title{\method: 3rd Place Solution for the MUMU Track of the 8th LSVOS Challenge at ECCV 2026}
\titlerunning{\method: 3rd Place Solution for MUMU Track}

\author{Chengfeng Qiu\inst{1}\\ Kaifeng Wei\inst{1}}
\authorrunning{Chengfeng Qiu and Kaifeng Wei}
\institute{$^{1}$Netease YiDun AI Lab, Hangzhou, China\\
\email{qiuchengfeng@corp.netease.com}, \email{hzweikaifeng@corp.netease.com}}

\maketitle

\begin{abstract}
The MUMU track of the 8th Large-scale Video Object Segmentation (LSVOS) Challenge requires a
\emph{single} unified multimodal model to jointly solve image tagging (Task~A), open-vocabulary
object detection (Task~B), and English captioning (Task~C) under strict resource constraints
($\leq$0.5B parameters and $\leq$8\,GB peak GPU memory).
We present \method, a training-free submission built on Microsoft Florence-2-base
($\sim$231M parameters), combining caption keyword mapping with multi-scale flip
test-time augmentation.
Task~C uses the native \texttt{<DETAILED\_CAPTION>} pathway with length/token sanitization.
Task~A maps the same detailed caption into the official quality/scene/event vocabularies via
an expanded keyword lexicon with whole-word matching and a lightweight expand-hints stage.
Task~B runs Florence-2 open detection (\texttt{<OD>}) with multi-scale and horizontal-flip
test-time augmentation (TTA), followed by label-aware non-maximum suppression (NMS).
Without fine-tuning, the system improves our reproduced Florence-2 baseline from
\textbf{15.16} to a best public score of \textbf{16.4815}, and ranks \textbf{3rd} on the
final MUMU leaderboard.
\end{abstract}

\section{Introduction}
\label{sec:intro}
The Large-scale Video Object Segmentation (LSVOS) Challenge series has become a major
benchmark venue for advancing video segmentation under realistic conditions
\cite{ding2023mose,hong2023lvos,ding2025mosev2,lsvos2025report}.
The \textbf{8th LSVOS Challenge}, held in conjunction with \textbf{ECCV 2026}, features
\textbf{four tracks}:
\begin{itemize}
  \item \textbf{MUMU Track} (this report): Multi-task Unified Multimodal Understanding.
  Participants must deploy \emph{one} shared model to produce (i)~hierarchical image tags,
  (ii)~object detections with open-vocabulary labels, and (iii)~English captions, under
  tight compute limits. Evaluation is performed on a held-out still-image test set
  (1{,}038 images in the public phase we target).
  \item \textbf{Classic VOS Track}: Semi-supervised video object segmentation, typically
  evaluated on long, complex sequences such as LVOS~\cite{hong2023lvos} and
  MOSE~\cite{ding2023mose}, measuring temporal mask consistency given first-frame masks.
  \item \textbf{Referring VOS (RVOS) Track}: Language-conditioned video segmentation, commonly
  centered on motion-focused datasets such as MeViS~\cite{ding2023mevis}, requiring
  joint vision--language grounding.
  \item \textbf{Complex VOS Track} (MOSEv2-style settings~\cite{ding2025mosev2}):
  Stress-tests robustness to small objects, frequent appearance/disappearance, heavy
  occlusion, adverse weather/lighting, and other real-world factors.
\end{itemize}
Across tracks, LSVOS emphasizes \emph{generalization beyond curated short clips}: long-term
identity preservation, crowded scenes, and multimodal conditioning where applicable.
Datasets historically associated with LSVOS include LVOS~\cite{hong2023lvos},
MOSE / MOSEv2~\cite{ding2023mose,ding2025mosev2}, and MeViS~\cite{ding2023mevis}, while the
MUMU track introduces a unified tagging--detection--captioning protocol on challenge images
with an official Codabench evaluation pipeline.
Unlike classic VOS, MUMU requires a \emph{single} checkpoint to serve three heterogeneous
outputs. Training three separate specialists violates the unified-model rule; large MLLMs often
exceed the 0.5B / 8\,GB budgets. We therefore adopt a compact vision--language foundation
model---Florence-2-base~\cite{xiao2024florence2}---and invest effort in \emph{prompt routing},
\emph{deterministic post-processing}, and \emph{detection TTA}, rather than fine-tuning.
Our best submission ranks 3rd on the final MUMU leaderboard with Final=16.4815
(A=26.49, B=5.45, C=44.05).

\subsubsection*{Contributions}
(1)~A practical Florence-2 recipe for MUMU that maintains one backbone for all tasks.
(2)~Caption-driven Task~A tagging with expanded lexicon and expand-hints fusion.
(3)~Multi-scale + flip OD TTA with NMS for Task~B, yielding denser, more stable boxes.

\section{Related Work}
\label{sec:related}
\subsubsection*{Unified vision--language models.}
Florence-2~\cite{xiao2024florence2} formulates vision tasks as sequence generation with
task tokens (\texttt{<CAPTION>}, \texttt{<DETAILED\_CAPTION>}, \texttt{<OD>}, etc.), enabling
one set of weights to cover captioning and detection. Other unified models exist, but
Florence-2-base uniquely fits MUMU's size limit while exposing native OD and detailed
caption heads. MUMU adds strict parameter and memory budgets on top of the standard
tagging--detection--captioning protocol, making compactness a first-class design goal.

\subsubsection*{Tagging from language.}
Zero-shot / weakly supervised tagging often maps free-form text to a closed vocabulary via
lexicons or CLIP-style matching~\cite{radford2021clip}. We adopt a lexicon-based pipeline:
it is bit-stable across machines and produces tags guaranteed to lie inside the official
vocabulary. The main cost is coverage of rare scene/event categories.

\subsubsection*{TTA for detection.}
Flips and multi-scale inference are standard in object detection
\cite{liu2016ssd,lin2017feature}. Because Florence-2 OD often outputs boxes without
calibrated scores, we rely on geometric redundancy + NMS rather than score thresholding;
when scores are constant, weighted averaging degenerates to plain averaging, so geometric
NMS is the natural fusion rule.

\section{Method}
\label{sec:method}

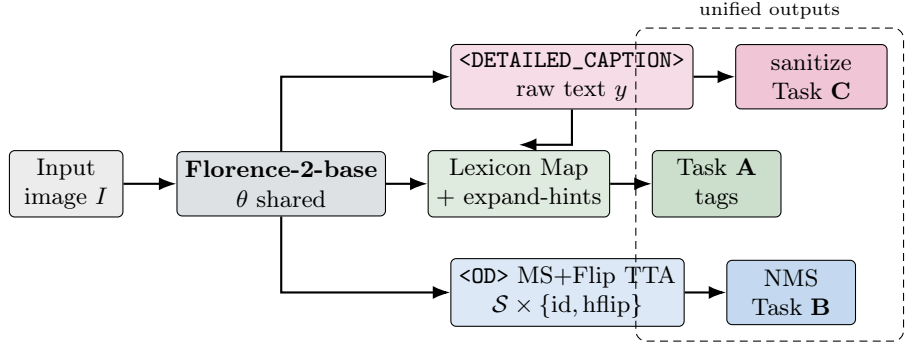
\begin{figure*}[t]
\centering
\begin{tikzpicture}[
  font=\footnotesize, >=Latex,
  node distance=0.35cm and 0.45cm,
  box/.style={draw, rounded corners=2pt, align=center, inner sep=3pt, minimum height=0.85cm},
  arr/.style={-{Latex}, thick}
]
\node[box, fill=gray!15, minimum width=1.5cm] (img) {Input\\image $I$};
\node[box, fill=boxM!18, minimum width=2.4cm, right=0.7cm of img] (f2) {\textbf{Florence-2-base}\\$\theta$ shared};
\node[box, fill=boxC!15, minimum width=2.3cm, above right=0.55cm and 0.85cm of f2] (cap) {\texttt{<DETAILED\_CAPTION>}\\raw text $y$};
\node[box, fill=boxC!25, minimum width=2.0cm, right=0.55cm of cap] (tc) {sanitize\\Task~\textbf{C}};
\node[box, fill=boxA!15, minimum width=2.3cm, right=0.55cm of f2] (map) {Lexicon Map\\$+$ expand-hints};
\node[box, fill=boxA!25, minimum width=1.7cm, right=0.55cm of map] (ta) {Task~\textbf{A}\\tags};
\node[box, fill=boxB!15, minimum width=2.6cm, below right=0.55cm and 0.85cm of f2] (od) {\texttt{<OD>} MS$+$Flip TTA\\$\mathcal{S}\times\{\mathrm{id},\mathrm{hflip}\}$};
\node[box, fill=boxB!25, minimum width=1.7cm, right=0.55cm of od] (tb) {NMS\\Task~\textbf{B}};
\draw[arr] (img) -- (f2);
\draw[arr] (f2) |- (cap);
\draw[arr] (cap) -- (tc);
\draw[arr] (f2) -- (map);
\draw[arr] (cap.south) |- ([yshift=2pt]map.north);
\draw[arr] (map) -- (ta);
\draw[arr] (f2) |- (od);
\draw[arr] (od) -- (tb);
\node[draw, densely dashed, rounded corners, fit=(tc)(ta)(tb), inner sep=6pt, label={[font=\scriptsize, anchor=south]above:unified outputs}] {};
\end{tikzpicture}
\caption{Overview of \method. One Florence-2-base checkpoint $\theta$ serves all MUMU tasks.
Task~A/C share the detailed-caption pathway; Task~B uses multi-view OD TTA + NMS.}
\label{fig:overview}
\end{figure*}

\subsection{Unified Backbone}
We use \texttt{microsoft/Florence-2-base}~\cite{xiao2024florence2} with
$|\theta| \approx 231.41\times 10^{6} \leq 0.5\times 10^{9}$ and peak memory
$M_{\mathrm{peak}}\approx 0.6\,\mathrm{GB}$ at batch size 1 ($\leq 8\,\mathrm{GB}$).
All tasks share $\theta$; we do \emph{not} fine-tune for the final submission.
Florence-2 casts each task as conditional sequence generation
$o = \mathrm{Decode}(f_\theta(I,\tau))$, where $\tau$ is a task token.
This choice is driven by three practical considerations: ample headroom under the 0.5B cap,
native task tokens for both detailed captioning and open detection (one checkpoint, no
task-specific heads), and peak memory well under 8\,GB so the full pipeline runs on a
single consumer GPU.

\subsection{Task~C: Detailed Captioning}
For each image $I$, we set $\tau=\texttt{<DETAILED\_CAPTION>}$ and obtain raw text $y$.
The submitted caption is
\begin{equation}
  c = \Pi_{\leq 30}\!\left(\mathrm{Trunc}_{\leq 300}\!\left(\mathrm{Norm}(y)\right)\right),
  \label{eq:sanitize}
\end{equation}
where $\mathrm{Norm}$ collapses whitespace, $\mathrm{Trunc}_{\leq 300}$ enforces the character
cap, and $\Pi_{\leq 30}$ greedily drops trailing whitespace-separated evaluator tokens until
the token count is $\leq 30$. Decoding uses beam search ($\mathrm{num\_beams}=3$), with
$\mathrm{batch\_size}=1$ fixed: padding in larger batches changes Florence generations and
breaks reproducibility. The pipeline is intentionally simple and deterministic. Florence's
detailed captions are often 50--80 words long, far exceeding the 30-token budget, so
truncation is unavoidable; we preserve the leading subject and setting and accept the loss
of trailing attribute lists.

\subsection{Task~A: Caption Keyword Mapping + Expand-Hints}
MUMU Task~A requires three tag lists from closed vocabularies
$\mathcal{V}_q$ (quality), $\mathcal{V}_s$ (scene), and $\mathcal{V}_e$ (event);
let $\mathcal{V}=\mathcal{V}_q\cup\mathcal{V}_s\cup\mathcal{V}_e$.

\subsubsection*{Lexicon matching.}
We maintain a phrase$\rightarrow$tag dictionary
$\mathcal{M}:\mathrm{phrase}\mapsto t\in\mathcal{V}$ (e.g., ``shopping mall''$\mapsto$
\texttt{shopping\_mall}). For $\tilde{y}=\mathrm{lower}(y)$, the hit set is
\begin{equation}
  \mathrm{Map}(y;\mathcal{M})
  =
  \left\{
    \mathcal{M}(p)
    \;\middle|\;
    p\in\mathrm{dom}(\mathcal{M}),\;
    p \text{ occurs in } \tilde{y} \text{ as a whole phrase}
  \right\},
  \label{eq:map}
\end{equation}
i.e., whole-word / whole-phrase matching. This avoids substring pitfalls such as
$\texttt{ski}\subset\texttt{skin}$ and $\texttt{mall}\subset\texttt{small}$; naive
substring matching produced hundreds of false scene tags per thousand images in early
versions of our lexicon.

\subsubsection*{Image-stat quality priors.}
Let $G$ be the grayscale image and $\Delta$ a discrete Laplacian operator. We compute
$v=\mathrm{Var}(\Delta G)$ and $\mu=\mathrm{Mean}(G)$ and produce heuristic quality tags
$Q(v,\mu)\subseteq\mathcal{V}_q$ (blur / low-light / under-/over-exposure thresholds),
used as a fallback when caption cues are weak.

\subsubsection*{Expand-hints fusion.}
Starting from a base-stage tag set $T_0$ (same caption pathway),
\begin{equation}
  T \;=\; T_0 \,\cup\, \mathrm{Map}(y;\mathcal{M}) \,\cup\, Q(v,\mu),
  \label{eq:expand}
\end{equation}
i.e., we only \emph{add} tags, then apply indoor/outdoor cleanup:
\begin{equation}
  \begin{aligned}
  &\textbf{if }\{\texttt{indoor},\texttt{outdoor}\}\subseteq T_s
  &&\textbf{then } T_s \leftarrow T_s\setminus\{\texttt{outdoor}\},\\
  &\textbf{else if }T_s\cap\mathcal{I}\neq\emptyset
  &&\textbf{then } T_s \leftarrow T_s\cup\{\texttt{indoor}\},\\
  &\textbf{else if }T_s\cap\mathcal{O}\neq\emptyset
  &&\textbf{then } T_s \leftarrow T_s\cup\{\texttt{outdoor}\},
  \end{aligned}
  \label{eq:inout}
\end{equation}
where $T_s=T\cap\mathcal{V}_s$ and $\mathcal{I},\mathcal{O}$ are indoor-/outdoor-indicative
scene subsets. The add-only rule guarantees expand-hints never reduces recall relative to
the base stage; the cleanup then removes mutually exclusive pairs.

\subsection{Task~B: Multi-Scale + Flip OD TTA}
Let $\mathcal{S}=\{0.75,1.0,1.25\}$ and $\mathrm{H}(\cdot)$ denote horizontal flip.
For scale $s$ with resized size $(W_s,H_s)$ and original size $(W,H)$, a box
$b_s=(x_1,y_1,x_2,y_2)$ is mapped back by
$\Phi_s(b_s)=(x_1\frac{W}{W_s},\,y_1\frac{H}{H_s},\,x_2\frac{W}{W_s},\,y_2\frac{H}{H_s})$,
and flip-space boxes are reflected before $\Phi_s$:
$\mathrm{H}^{-1}(b)=(W-x_2,\,y_1,\,W-x_1,\,y_2)$.
The proposal pool is
\begin{equation}
  \mathcal{P}
  =
  \bigcup_{s\in\mathcal{S}}
  \left(
    \Phi_s\!\left(\mathrm{OD}(I_s)\right)
    \;\cup\;
    \Phi_s\!\left(\mathrm{H}^{-1}(\mathrm{OD}(\mathrm{H}(I_s))))\right)
  \right).
  \label{eq:pool}
\end{equation}
Label-aware NMS keeps a box $b$ against kept set $\mathcal{K}$ iff
$\forall k\in\mathcal{K}$:
$\mathrm{label}(b)\neq\mathrm{label}(k)$
$\lor\ \mathrm{IoU}(b,k)\leq \tau$,
with $\tau=0.5$ and $|\mathcal{K}|\leq 300$; malformed labels
(commas\slash colons\slash length${>}60$) are discarded for schema validity.
The three scales balance coverage and cost: smaller scales recover small objects missed at
native resolution, larger scales help crowded scenes, and the flip view adds complementary
proposals for asymmetric objects.
Florence OD often returns empty score lists; we then set $s(b)\equiv 0.85$. With constant
scores thresholding cannot rank boxes, so geometric TTA + NMS is our main source of gain.

\subsection{Inference Recipe and Design Principles}
Putting everything together: (i)~$y \leftarrow$ caption with
$\tau=\texttt{<DETAILED\_CAPTION>}$, $c\leftarrow$ Eq.~\eqref{eq:sanitize};
(ii)~$T \leftarrow$ Eqs.~\eqref{eq:map}--\eqref{eq:inout} (Task~A);
(iii)~$\mathcal{K} \leftarrow$ NMS on $\mathcal{P}$ from Eq.~\eqref{eq:pool} (Task~B);
(iv)~emit $\{T,\mathcal{K},c\}$ with official \texttt{model\_info} fields only
(\texttt{parameters\_m}, \texttt{gflops\_224}, \texttt{peak\_memory\_gb}).
Three principles guided all changes: \emph{do not hurt Task~C} (captioning dominates the
Final score, so its path is fixed once matching the 15.16 run); \emph{prefer deterministic
post-processing over fine-tuning}; and \emph{spend test-time compute where scores improve}.

\section{Experiments}
\label{sec:exp}
\subsubsection*{Setup.}
Public MUMU test split: \textbf{1{,}038} images. We do not use private labels; all ablations
use online Codabench scores or offline prediction statistics. Our reproduced baseline runs
the official inference script unchanged (\texttt{<DETAILED\_CAPTION>} for C, official tag
mapping for A, single-view \texttt{<OD>} for B), yielding Final=15.1568
(A=21.15, B=4.77, C=44.05). Task~C is evaluated with CIDEr-D, SPICE, and CLIPScore; since
C dominates the Final score, we preserve caption quality and mainly improve A/B.
Implementation: PyTorch + HuggingFace Transformers, \texttt{trust\_remote\_code=True},
FP16 on a single consumer GPU; full test inference with multi-scale TTA takes roughly
50--70 minutes at $\sim$0.25 img/s.

\subsubsection*{Main results.}
Table~\ref{tab:main} summarizes the progression. Expand-hints + flip TTA lifts Final to
16.4496 (+1.29), with A $\approx$+5.3 and B $\approx$+0.6 while C is unchanged; multi-scale
TTA further improves Final to \textbf{16.4815} via denser and more accurate boxes. Our best
Task~C run achieves CIDEr-D=41.26, SPICE=16.31, CLIPScore=75.52.

\begin{table}[t]
\centering
\caption{Public Codabench scores on the MUMU test set (1{,}038 images).
``Hints'' = expand-hints for Task~A; ``Flip'' = horizontal-flip OD TTA;
``MS'' = multi-scale OD TTA $\{0.75,1.0,1.25\}$. Row 3 is the final leaderboard entry
with Final=16.4815.}
\label{tab:main}
\setlength{\tabcolsep}{4pt}
\begin{tabular}{lccc}
\toprule
Method & Final & A / B / C \\
\midrule
Florence-2-base, bs=1 (baseline) & 15.1568 & 21.15 / 4.77 / 44.05 \\
+ Hints + Flip TTA & 16.4496 & 26.49 / 5.34 / 44.05 \\
+ Hints + MS+Flip TTA (best) & \textbf{16.4815} & 26.49 / 5.45 / 44.05 \\
\bottomrule
\end{tabular}
\end{table}

\subsubsection*{Task analysis.}
\emph{Task~A}: expand-hints lifts A from 21.15 to 26.49 (+5.34), almost entirely from
recall; precision is preserved because we only add tags. \emph{Task~B}: flip TTA raises B
from 4.77 to 5.34 (+0.57) by recovering boxes missed under horizontal asymmetry; adding
multi-scale TTA increases average boxes per image from 9.93 to 11.44 and reduces empty
detections from 17 to 16 (Table~\ref{tab:offline}), giving the final 5.45.
\emph{Task~C}: stays at 44.05 across all rows by design---the sanitization pipeline is
applied identically and the \texttt{<DETAILED\_CAPTION>} path is never modified, which
isolates the A/B contributions.

\begin{table}[t]
\centering
\caption{Offline Task~B statistics on the full test set, same A/C as the 16.4815 submission.}
\label{tab:offline}
\begin{tabular}{lccc}
\toprule
Detection recipe & Avg.\ \#boxes & Empty & Max \\
\midrule
Single-view OD (baseline) & 8.91 & 19 & 74 \\
+ Flip TTA + NMS@0.5 & 9.93 & 17 & 84 \\
+ MS$\{0.75,1,1.25\}$ + Flip + NMS@0.5 & \textbf{11.44} & \textbf{16} & 92 \\
\bottomrule
\end{tabular}
\end{table}

\subsubsection*{Ablations and negative results.}
\emph{Batch size}: bs$>$1 changes captions due to padding and breaks reproducibility of the
15.16 baseline, so we fix bs=1. \emph{Florence-2-base-ft}: the fine-tuned checkpoint
underperformed ($\sim$8 Final), likely due to distribution shift and shorter captions
hurting C. \emph{Score thresholds}: IoU sweeps at $\{0.4,0.5,0.6\}$ change flip-only
results by less than 0.1 Final. \emph{PromptGen / Danbooru taggers}: shorter captions
(hurting C) and non-MUMU vocabularies; after remapping they still cannot outperform
\method. \emph{Substring lexicon bugs}: early substring matching produced false tags such
as \texttt{snow} from ``skin'' and \texttt{shopping\_mall} from ``small''; whole-phrase
matching plus cleanup was the single most impactful fix for Task~A precision.

\subsubsection*{Error analysis.}
\emph{Task~A}: failures come from captions missing scene nouns (close-up portraits, product
shots) and events requiring world knowledge. \emph{Task~B}: empty detections persist on
textureless landscapes and abstract images where no clear object boundary exists.
\emph{Task~C}: the 30-token limit forces aggressive truncation; a learned ranker selecting
the best 30-token subsequence could recover some lost attribute information.

\section{Discussion and Conclusion}
\label{sec:disc}
MUMU rewards systems that are \emph{small, unified, and carefully post-processed}. Our gains
come almost entirely from (i)~better use of Florence captions for closed-vocabulary tagging
and (ii)~extra test-time compute for detection. Remaining headroom lies in Task~C (better
selection under the 30-token limit) and calibrated detection scores for smarter box fusion.
Our system has three limitations: Task~C is capped by the 30-token budget; Task~B still
produces empty detections on textureless imagery; and the Task~A lexicon is hand-curated and
may not cover rare categories in future phases. Promising future directions include
fine-tuning Florence-2 within the 0.5B/8\,GB envelope, learning a lightweight caption ranker,
and calibrating OD scores for score-based fusion.

We described \method, our 3rd-place submission to the MUMU Track of the 8th LSVOS Challenge
at ECCV 2026, achieving a final score of \textbf{16.4815}. Built on Florence-2-base, it
combines detailed captioning, lexicon-based tagging with expand-hints, and multi-scale flip
OD TTA under strict resource limits. The pipeline is training-free and reproducible, showing
that careful prompting and TTA can be strong tools when model size is capped.

\subsubsection*{Acknowledgements.}
We thank the LSVOS Challenge organizers and Codabench maintainers for the evaluation platform
and baseline releases.


\end{document}